\documentclass{article}
\usepackage{spconf,amsmath,epsfig}
\usepackage{soul,color}
\usepackage{multirow}
\usepackage{float}
\usepackage{url}

\title{Visualization of Deep Transfer Learning In SAR Imagery}
\name{Abu Md Niamul Taufique, Navya Nagananda, and Andreas Savakis
\thanks{This research was supported in part 
by the Air Force Office of Scientific Research (AFOSR) under Dynamic Data Driven Applications Systems (DDDAS) grant FA9550-18-1-0121.}}
\address{Rochester Institute of Technology}
\begin{document}
%
\maketitle

\begin{abstract}
Synthetic Aperture Radar (SAR) imagery has diverse applications in land and marine surveillance. Unlike electro-optical (EO) systems, these systems are not affected by weather conditions and can be used in the day and night times. With the growing importance of SAR imagery, it would be desirable if models trained on widely available EO datasets can also be used for SAR images. In this work, we consider transfer learning to leverage deep features from a network trained on an EO ships dataset and generate predictions on SAR imagery. Furthermore, by exploring the network activations in the form of class-activation maps (CAMs), we visualize the transfer learning process to SAR imagery and gain insight on how a deep network interprets a new modality. 
\end{abstract}
\begin{keywords}
SAR, Transfer Learning, Explainability, Class Activation Maps.
\end{keywords}

\section{Introduction}
\begin{figure}[t]
\centerline{\includegraphics[width = 0.5\textwidth]{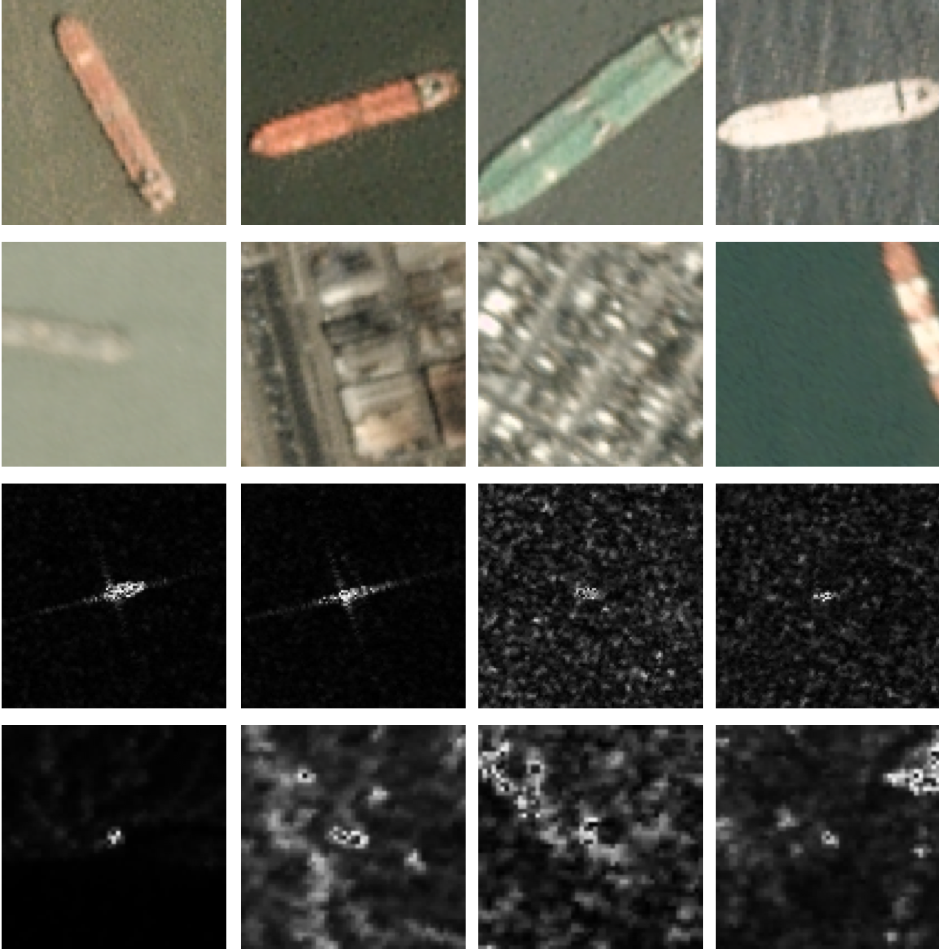}}
\caption{Examples from the EO ships and SAR ships datasets. The first and second rows show positive and negative samples of EO ships respectively. 
The third and fourth row show positive and negative samples of SAR ships respectively.}
\label{fig:dataset}
\end{figure}

Remote sensing imagery has been growing in volume  and significance in recent years. 
The deployment of commercial satellites, by Planet Labs and Digital Globe, is making satellite imagery accessible at an unprecedented scale, which requires automated methods for analysis and interpretation. 
Deep learning techniques have been successfully used for a variety of remote sensing scenarios.
However, Synthetic Aperture Radar (SAR) images are less accessible and harder to label than standard Electro-Optical (EO) data. 
Fig. \ref{fig:dataset} shows examples of EO and SAR imagery. These examples illustrate the visual differences between the two modalities, and the challenge of using networks pretrained on EO for SAR image classification.  
Due to these limitations, training deep networks with SAR data is much more challenging. 

In this paper, we utilize EO data for pretraining a deep network and then perform transfer learning with SAR data to the new domain. 
We then examine the network activations, before and after transfer learning, from EO to SAR, to gain insights into the transfer learning process across modalities. 

\section{Related Work}

\begin{figure}[t]
\centerline{\includegraphics[width = 0.35\textwidth]{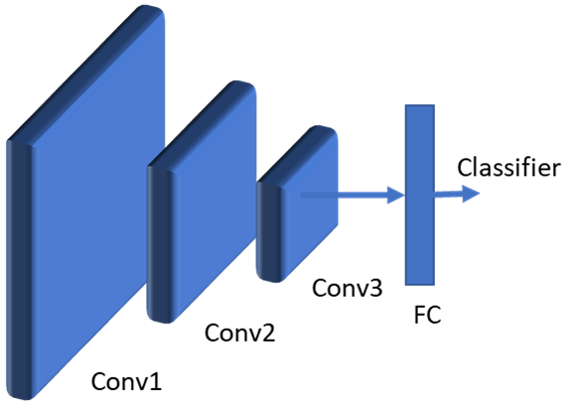}}
\caption{Network architecture used for training on EO data and transfer learning.}
\label{fig:network}
\end{figure}

The objective of transfer learning is to transfer knowledge, from a source domain where it is easy to obtain labeled data for training, to a target domain where data is scarce \cite{TransferLearning2014}.
Pretraining in the source domain helps generate robust features in the network and 
reduces the amount of labeled data required for training our model \cite{Rostami_2019}. 
In our scenario, the source domain contains EO data, which is labeled and readily available. 

Explainability is important to interpret how the deep network makes decisions.
To examine how a deep network interprets a scene, we use Class Activation Mapping (CAM) techniques that visualize the network's internal representation. Aerial CAM~\cite{ivmsp2018} is a method that is used to obtain the CAM to understand the networks perception and to identify salient regions of the scene. 
It makes use of the Global Average Pooling (GAP) layer to obtain the CAM. The GAP layer is a weighted sum of the feature maps from the last convolutional layer the provides the class activation heat maps for each predicted class~\cite{ivmsp2018}, which are thresholded to produce a mask that displays the highest activating region of the image.

The Gradient-weighted Class Activation Mapping (Grad-CAM)~\cite{GradCam2016} is a more general technique based on the gradient. Grad-CAM makes use of the target label information available at the final convolution layer to produce a coarse localization map that highlights the salient regions of the images used to produce that particular target label. Grad-CAM is applicable to wide variety of CNNs and does not require the GAP layer. 
In this work, we employ Grad-CAM to visualize the salient regions before and after transfer learning from the EO to the SAR domain.

\section{Methodology}
We designed a small convolutional network \cite{AlexNet}, suitable for the size of the SAR dataset considered, to study  transfer learning from EO to SAR. The network architecture is shown in Fig. \ref{fig:network}. 
The network consists of three convolutional layers and a fully connected layer. 
A max pooling layer  and ReLU activation are used between Conv1 and Conv2.
In between the Conv2 and Conv3 layers, dropout, max pooling, and ReLU activation are applied.
A ReLU activation layer is applied after the Conv3 layer and a global average pooling layer is used before the fully connected layer.

\subsection{Datasets and Experiments}
In our experiments, we used the dataset "EO Ships in Satellite Imagery" \cite{EOships}, and the SAR ships dataset \cite{SAR:Schwegmann2016}. 
The images provided in the EO ships dataset are taken using Planet satellites over the San Francisco Bay and San Pedro Bay areas of California.
The dataset contains 1000 chips with ships and 3000 chips without ships. Example chips are shown in Fig. \ref{fig:dataset}.  

\begin{table}[ht]
\caption{Attributes in the SAR Ships test samples. } 
\label{tab:attributes}
\begin{center}        
\begin{tabular}{c|c|c|c} 
\hline\hline
\multicolumn{2}{c|}{Attributes} & Ships & No ships \\
\hline\hline
\multirow{3}{*}{Sensor} & GRDH & 95 & 1756 \\
& GRDM & 207 & 4116 \\
& SCNA & 18 & 512  \\
\hline
\multirow{4}{*}{Polarization} & HH & 46 & 1124 \\
& HV & 35 & 554  \\
& VV & 106 & 2699  \\
& VH & 133 & 2007 \\
\hline
\multirow{3}{*}{Incidence Angle} & Small & 41 & 1735  \\
& Medium & 102 & 2269 \\
& Large & 177 & 2380  \\
\hline
\hline\hline
\end{tabular}
\end{center}
\end{table}

\begin{table*}[h]
\caption{Performance comparison between EO trained model on SAR before and after performing Transfer Learning (TL).
} 
\label{tab:scores}
\begin{center}        
\begin{tabular}{c|c|ccc|cccc|ccc|c} 
\hline\hline
\multicolumn{2}{c|}{} & \multicolumn{3}{c|}{Sensor} & \multicolumn{4}{c|}{Polarization} & \multicolumn{3}{c|}{Incidence Angle} & \\
\hline
Training & Class & GRDH & GRDM & SCNA & HH & HV & VV & VH & Small & Medium & Large & Overall\\
\hline\hline
\multirow{3}{*}{Only EO} & Ship & 0.34 & 0.19 & 0.83 & 0.74 & 0.06 & 0.47 & 0 & 0.15 & 0.22 & 0.33 & 0.27 \\
& No ship & 0.75 & 0.71 & 0.62 & 0.62 & 0.93 & 0.54 & 0.95 & 0.57 & 0.73 & 0.81 & 0.72 \\
& Overall & 0.55 & 0.45 & 0.73 & 0.68 & 0.49 & 0.51 & 0.48 & 0.36 & 0.47 & 0.57 & 0.49 \\
\hline
\multirow{3}{*}{TL to SAR} & Ship &  0.89 & 0.92 & 0.83 & 0.76 & 0.94 & 0.93 & 0.93 & 0.9 & 0.91 & 0.91 & 0.91 \\
& No ship & 0.99 & 0.94 & 0.96 & 0.98 & 0.97 & 0.97 & 0.91 & 0.96 & 0.95 & 0.94 & 0.95 \\
& Overall & 0.94 & 0.93 & 0.9 & 0.87 & 0.96 & 0.95 & 0.92 & 0.93 & 0.93 & 0.93 & 0.93 \\
\hline\hline

\end{tabular}
\end{center}
\end{table*}

The SAR ships data contains Sintel-1 Extended Wide Swath images and RADARSAT-2 ScanSAR images
collected in
2014 and 2015 at the South African Exclusive Economic Zone.  
The SAR dataset contains 1596 positive samples, containing a ship, and 7980 negative samples that do not contain a ship. Among the positive samples, we randomly selected 80\% for training and 20\% for testing. Among the negative samples, we randomly selected 20\% for training and 80\% for testing to keep the balance between positive and negative samples in the training set.   

Various attributes are annotated in the dataset which describe unique conditions for the SAR image acquisition.
We categorized the dataset based on the sensor resolution, polarization, and incidence angle of the radar source signal. We only use these attributes for categorizing the test samples, as outlined in Table \ref{tab:attributes}.
The Sentinel-1 has two types of data available, which are Ground Range Detected (GRD) imagery with High (GRDH) and Medium (GRDM) resolutions. The resolution of GRDH is 50x50 $m^2$ with 25x25 $m^2$ pixel spacing and the resolution of GRDM is 93x97 $m^2$ with 40x40 $m^2$ pixel spacing. The RADARSAT-2 satellite has three ScanSAR Narrow (SCNA) imagery available with 81x30 $m^2$ resolution and 25x25 $m^2$ pixel spacing. 
There are four types of polarization state possible based on the electromagnetic wave of the transmitted signal namely HH, HV, VV, and VH polarization. 
A total of 43 Sentinel-1 images are available among SDH (HH+HV), SDV (VV+VH), and SSV (VV) Extra Wide (EW) polarization acquisition modes. No GRDM HH, HV and SCNA HV, VV, or VH images are present in the dataset. 
Another important attribute for the SAR dataset acquisition is the incidence angle, which is in the range of $19.0 - 47.0$ and $20.0 - 39.0$ for Sentinel-1 and RADARSAT-2 sensors, respectively. 
We categorized the dataset into three categories based on the incidence angle, i.e. small, medium, and large incidence angles with the range of (19.0,25.0], (25.0,35.0], and (35.0,47.0] degrees, respectively. 

\begin{figure}[ht]
\centerline{\includegraphics[width = 0.45\textwidth]{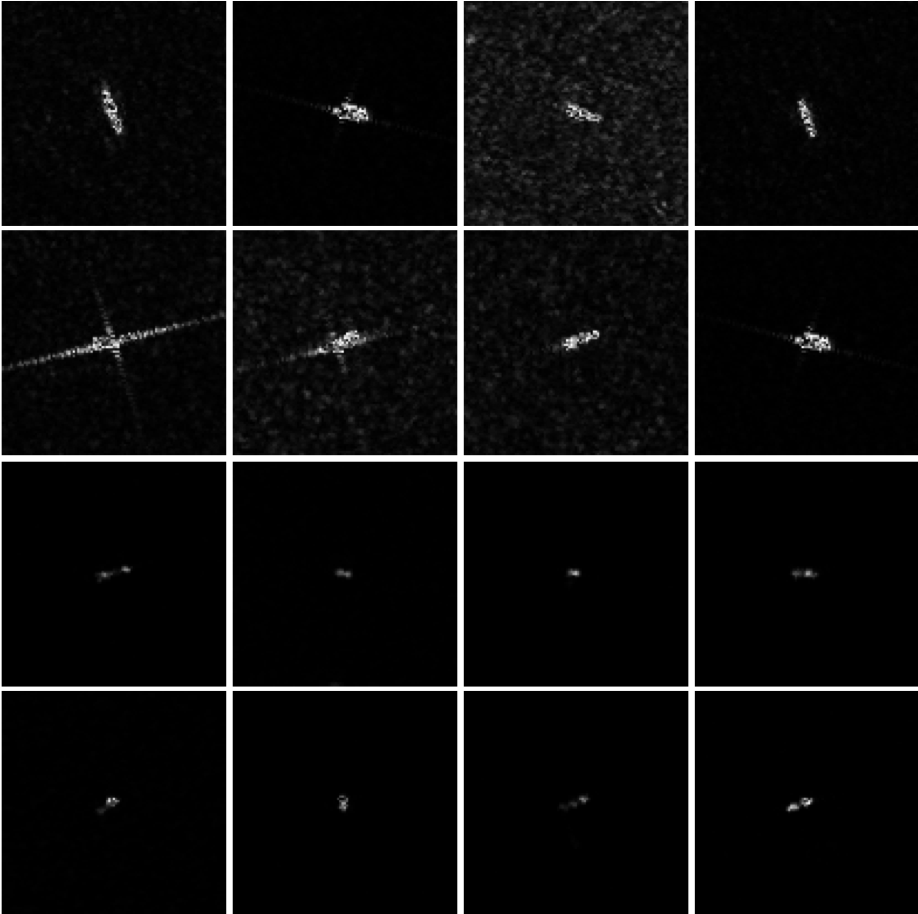}}
\caption{Ship samples showing significant difference in ship size and brightness among 
SCNA, HH, VH, and HV
attributes, shown from top to bottom rows, respectively.}
\label{fig:network}
\end{figure}

In our experiments, we initially trained our network on the EO dataset with Adam optimizer \cite{kingma2014adam} and cross-entropy loss. The network was trained with a learning rate of 0.0001 for 100 epochs.
This model was then used to test the SAR dataset before and after transfer learning.
For transfer learning, we initialize the model weights with those obtained from  training the model with EO data and finetune all the layers except the first convolutional layer. The network is finetuned using the Adam optimizer with 0.0001 learning rate and categorical cross-entropy loss.

\section{Results}

The normalized mean per class accuracy results are presented in Table \ref{tab:scores}. Overall, the only EO training without transfer learning does not achieve good performance, given that the EO dataset and the SAR dataset have significantly different distributions.
The classification results for the RADARSAT-2 SCNA, HH polarized, VH polarized, and HV polarized images chips are quite different.
Inspection of ship samples from these categories revealed that
most of the SCNA and HH polarized ships are bright and easily distinguishable from the background, while most of the VH and HV polarized ships are small in size. Due to this difference, SCNA and HH polarized ships are better recognized by the network with just the EO training, but no VH polarized ships and few HV polarized ships are recognized. Overall, transfer learning to the SAR domain significantly improved the discriminative capability of the network.  

\begin{figure}[h]
\centerline{\includegraphics[width = 0.5\textwidth]{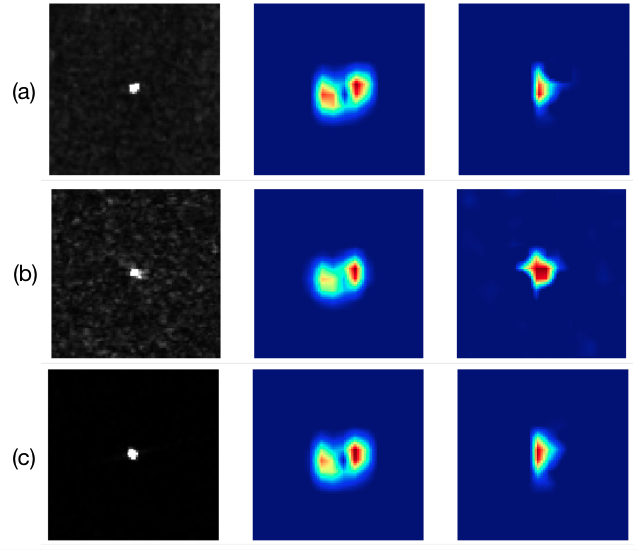}}
\caption{Visualization of CAMs for SAR images (left) using EO trained model (middle) and after transfer learning to SAR (right). The image is classified correctly both before and after transfer transfer learning. Image (a) is from GRDM with VV polarization, (b) is from SCNA with HH polarization, and (c) is from GRDH with HH polarization.}
\label{fig:vis1}
\end{figure}

To illustrate the transfer learning process, we obtained Grad-CAM visualizations of our network trained on EO and tested on SAR imagery with transfer learning.
The CAMs for only those ships which were correctly classified in the EO and SAR domains are shown in Fig. \ref{fig:vis1}. From these maps we can see that the network starts to focus more on the location of the actual ship after transfer learning.

\begin{figure}[h]
\centerline{\includegraphics[width = 0.5\textwidth]{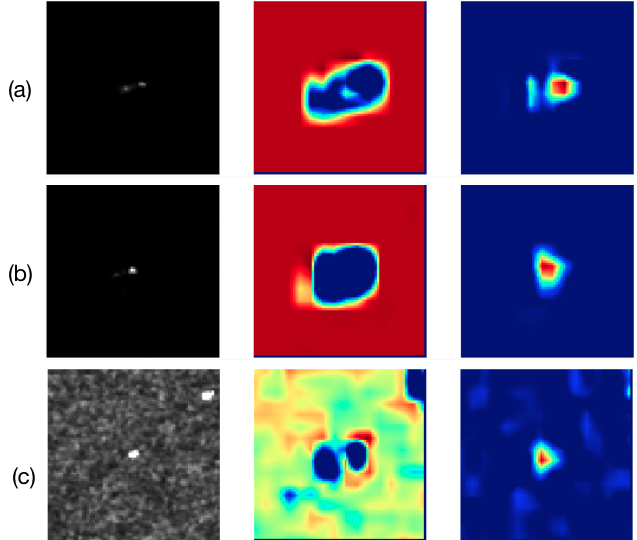}}
\caption{Visualization of CAMs for SAR images (left) using EO trained model (middle) and after transfer learning to SAR (right). The image is incorrectly classified with only EO training and correctly classified after transfer learning to SAR. Image (a) is from GRDH with VH polarizarion, (b) is from GRDH with HV polarization, and (c) is from GRDM with VV polarization.}
\label{fig:vis2}
\end{figure}

For the CAM heatmaps generated for correct classification 
before transfer learning, shown in the middle column of Fig. \ref{fig:vis1}, the peaks are less pronounced, and the general region around the object is selected as the salient region for classification. 
After transfer learning, in the heatmaps shown in the right column of Fig. \ref{fig:vis1}, the salient region is more pronounced and centered on the target,
with lesser emphasis on the surrounding region.


From the heat maps in the middle column of Fig. \ref{fig:vis2}, generated 
using the network trained on EO before transfer learning, 
the regions selected for classification are not on the ship but rather on surrounding spurious areas resulting in the wrong classification.
After transfer learning, however, the salient region is shifted towards the object, as shown in the right column of Fig. \ref{fig:vis2},  thus providing the correct classification.



\section{Conclusions}
In this work we demonstrated the effectiveness of deep transfer learning from EO imagery to SAR imagery using Grad-CAM network visualization technique. 
We categorized the test dataset based on some significant attributes annotated in the dataset such as the type of sensor resolution, polarization, and incidence angle used for image acquisition to further understand the performance variation based on various attributes of the dataset. 

The classification results showed interesting results where for RADARSAT-2 and HH polarized imagery, ships are well recognized with only EO training. It is also interesting that no ships were detected in VH polarized imagery and a few shps were detected in HV polarized imagery with only EO training but performance improved significantly with transfer learning.
We attribute this behaviour to the fact that in the RADARSAT-2 imagery, ships are bright and well distinguishable from the background where the VH polarized ships are tiny in size. 
In general, transfer learning improved the classification performance significantly. 
The further demonstration using Grad-CAM visualization shows us the effectiveness of deep transfer learning from EO to SAR domain.   
\bibliographystyle{IEEEbib}
\bibliography{refs}

\begin{thebibliography}{1}

\bibitem{TransferLearning2014}
M.~{Oquab}, L.~{Bottou}, I.~{Laptev}, and J.~{Sivic},
\newblock ``Learning and transferring mid-level image representations using
  convolutional neural networks,''
\newblock in {\em Computer Vision and Pattern Recognition}, June 2014, pp.
  1717--1724.

\bibitem{Rostami_2019}
Mohammad Rostami, Soheil Kolouri, Eric Eaton, and Kyungnam Kim,
\newblock ``Deep transfer learning for few-shot sar image classification,''
\newblock {\em Remote Sensing}, vol. 11, no. 11, pp. 1374, Jun 2019.

\bibitem{ivmsp2018}
B.~Vasu, F.~U. Rahman, and A.~Savakis,
\newblock ``Aerial-cam: Salient structures and textures in network class
  activation maps of aerial imagery,''
\newblock in {\em 2018 IEEE 13th Image, Video, and Multidimensional Signal
  Processing Workshop (IVMSP)}, 2018.

\bibitem{GradCam2016}
R.~R. {Selvaraju}, M.~{Cogswell}, A.~{Das}, R.~{Vedantam}, D.~{Parikh}, and
  D.~{Batra},
\newblock ``Grad-cam: Visual explanations from deep networks via gradient-based
  localization,''
\newblock in {\em 2017 IEEE International Conference on Computer Vision
  (ICCV)}, Oct 2017, pp. 618--626.

\bibitem{AlexNet}
Alex Krizhevsky, Ilya Sutskever, and Geoffrey~E Hinton,
\newblock ``Imagenet classification with deep convolutional neural networks,''
\newblock in {\em Advances in neural information processing systems}, 2012, pp.
  1097--1105.

\bibitem{EOships}
R.~Hammell,
\newblock ``Ships in satellite imagery, ver.~9,'' \url{https:
  //www.kaggle.com/rhammell/ships-in-satellite-imagery}, 2018.

\bibitem{SAR:Schwegmann2016}
C.~P. Schwegmann, W.~Kleynhans, B.~P. Salmon, L.~W. Mdakane, and R.~G.~V.
  Meyer,
\newblock ``{Very deep learning for ship discrimination in Synthetic Aperture
  Radar imagery},''
\newblock in {\em {IEEE International Geoscience and Remote Sensing Symposium
  (IGARSS)}}, July 2016, pp. 104--107.

\bibitem{kingma2014adam}
Diederik~P Kingma and Jimmy Ba,
\newblock ``Adam: A method for stochastic optimization,''
\newblock {\em arXiv preprint arXiv:1412.6980}, 2014.

\end{thebibliography}

\end{document}